\documentclass[conference]{IEEEtran}
\IEEEoverridecommandlockouts
\usepackage{cite}
\usepackage{amsmath,amssymb,amsfonts}
\usepackage{algorithmic}
\usepackage{booktabs}
\usepackage{times}
\usepackage{tikz}
\usepackage{pgf-pie}
\usepackage{graphicx}
\usepackage[most]{tcolorbox}
\usepackage{xparse}
\usepackage{xspace}
\usepackage{enumitem}
\usepackage{textcomp}
\usepackage{xcolor}
\usepackage{hyperref}
\usepackage{float}
\usepackage{multirow}
\usepackage{tabularx}
\usepackage{lipsum}
\usepackage{subfigure}

\makeatletter
\newcommand{\stitle}[1]{\smallskip\noindent\textbf{#1\@addpunct{}}}
\newcommand{\ititle}[1]{\smallskip\noindent\emph{#1\@addpunct{}}}
\makeatother

\definecolor{brown-web}{rgb}{0.65, 0.16, 0.16}
\newcommand{\blue}[1]{\textcolor{blue}{#1}}
\newcommand{\ex}[1]{\textcolor{brown-web}{#1}}

\newcommand\copyrighttext{%
  \footnotesize \textcopyright 2024 IEEE. Personal use of this material is permitted.
  Permission from IEEE must be obtained for all other uses, in any current or future
  media, including reprinting/republishing this material for advertising or promotional
  purposes, creating new collective works, for resale or redistribution to servers or
  lists, or reuse of any copyrighted component of this work in other works.
  }
  
\makeatletter
\newcommand{\confname}[1]{%
\def\ps@IEEEtitlepagestyle{
\def\@oddhead{\parbox{\textwidth}{\@IEEEheaderstyle #1}}%
\let\@evenhead\@empty
\def\@oddfoot{\parbox{\textwidth}{\mycopyrightnotice}}%
\let\@evenfoot\@empty}}
\newcommand{\mycopyrightnotice}{%
\fbox{\parbox{\dimexpr\textwidth-\fboxsep-\fboxrule\relax}{\copyrighttext}}
}

\confname{\textbf{\ex{DRAFT COPY}}: Accepted to be Published in 2024 27th International Conference on Computer and Information Technology (ICCIT)
}

\begin{document}

\title{BN-AuthProf: Benchmarking Machine Learning for Bangla Author Profiling on Social Media Texts}

\author{
\IEEEauthorblockN{Raisa Tasnim}
\IEEEauthorblockA{
    \textit{Dept. of CSE}, \textit{Premier University}\\
    Chittagong, Bangladesh\\
    raisatasnim14@gmail.com
    }
\vspace{-3mm}
\and
\IEEEauthorblockN{Mehanaz Chowdhury}
\IEEEauthorblockA{
    \textit{Dept. of CSE}, \textit{Premier University}\\
    Chittagong, Bangladesh\\
    mehanaz.chy09@gmail.com
    }
\vspace{-3mm}
\and
\IEEEauthorblockN{Md Ataur Rahman}
\IEEEauthorblockA{
    \textit{ESSI}, \textit{Universitat Polit\`{e}cnica de Catalunya}\\
    Barcelona, Spain\\
    md.ataur.rahman@upc.edu
    }
\vspace{-3mm}
}


\maketitle
\vspace{-4mm}

\begin{abstract}
Author profiling, the analysis of texts to uncover attributes such as gender and age of the author, has become essential with the widespread use of social media platforms. This paper focuses on author profiling in the Bangla language, aiming to extract valuable insights about anonymous authors based on their writing style on social media. The primary objective is to introduce and benchmark the performance of machine learning approaches on a newly created Bangla Author Profiling dataset, BN-AuthProf\footnote{\label{footnote-dataset}\texttt{\textbf{\ex{Code \& Artifacts}}}: \blue{\url{https://github.com/crusnic-corp/BN-AuthProf}}}.
The dataset comprises 30,131 social media posts from 300 authors, labeled by their age and gender. Authors' identities and sensitive information were anonymized to ensure privacy. Various classical machine learning and deep learning techniques were employed to evaluate the dataset. For gender classification, the best accuracy achieved was 80\% using Support Vector Machine (SVM), while a Multinomial Naive Bayes (MNB) classifier achieved the best F1 score of 0.756. For age classification, MNB attained a maximum accuracy score of 91\% with an F1 score of 0.905. This research highlights the effectiveness of machine learning in gender and age classification for Bangla author profiling, with practical implications spanning marketing, security, forensic linguistics, education, and criminal investigations, considering privacy and biases.
\end{abstract}

\medskip


\section{Introduction}

Author profiling has become indispensable with the growing usage of social media, where anyone can write anything \cite{Lundeqvist2017AuthorPA}. It is the study of specific texts to discover unique characteristics of an author, such as gender, age group, region, and language, based on their writing style and content \cite{Santosh2013AuthorPP}. As social media platforms expand, the need to unravel the characteristics of authors becomes increasingly apparent in order to ensure the safety of online activities and communication. Author profiling has significant implications for security, forensic linguistics, education, research, and marketing. For example, social media ads are now mostly curated based on author characteristics, tailoring content to different demographics. Companies are leveraging authors' profiles to understand reviews and better grasp their target audience. Forensic linguists utilize author profiling to determine the linguistic profile of suspects, aiding investigations \cite{Rangel2013AuthorPI}. Moreover, it could be useful in detecting fake profiles and news \cite{Zhang2017AnAF}. Nevertheless, discerning the identities behind texts is a complex and challenging task that needs further attention, especially for a low-resourced language as Bangla \cite{hossain2024authornet}.


\stitle{Problem Statement and Our Proposal:} While author profiling has been extensively explored in languages like Dutch, English, Greek, and Spanish \cite{stamatatos2014overview}, it remains largely unexplored in Bangla due to the unavailability of a benchmarked dataset. This paper aims to fill this gap by benchmarking classical machine learning (ML) and deep learning (DL) approaches for predicting authors' gender and age based on Bangla textual content from social media platforms. For this purpose, we extensively compiled the \textbf{BN-AuthProf Dataset}\textsuperscript{\ref{footnote-dataset}} by collecting author IDs with substantial Bangla posts with their consent. The dataset consists of 300 anonymized authors and 30,131 manually curated Facebook posts in Bangla, tagged with age and gender information.

\vspace{1mm}
The \textbf{contributions} of this paper are summarized as follows:
\begin{itemize}
    \setlength\itemsep{0.3em}
    \item We studied author profiling in the Bangla language, aiming to predict demographic factors such as age group and gender based on textual content. To the best of our knowledge, this is the first work of its kind in Bangla.
    
    \item We introduced the BN-AuthProf Dataset that includes 30,131 Facebook status from 300 Bengali authors, each labeled with age and gender. We also compared nine popular ML/DL algorithms for benchmarking.

    \item We have drawn several conclusions and recommendations for the research community to consider. Additionally, our results could serve as a strong baseline for future machine learning experiments in this domain.
\end{itemize}

\section{Related Work}
In the recent past, automatic profiling of authorship has seen a multitude of literary developments. Here we categorically present these key findings.

\stitle{Language Based:}
Though researchers have extensively explored approaches across various languages, they are still in their infancy when it comes to low-resource languages like Bangla \cite{ahmed2024emotion}. A lexical model was used in an early attempt to predict age and gender in English on social networks\cite{sap2014developing}. This system used regression and classification models and yielded results (91.9\% accuracy) in line with state-of-the-art age and gender prediction. Basile et al. \cite{basile2018simply} employed a gender prediction model across English, Spanish, Arabic, and Portuguese on Twitter data which performed between 0.68 and 0.98, with an average accuracy of 0.86 on the testset. For identifying Indianic native languages, TF-IDF and n-gram features showed promising results \cite{nayel2017mangalore}. Alsmear et al. \cite{alsmearat2014extensive} studied gender identification in Arabic texts, employing various classification techniques.

\stitle{Features Based:}
Peersman et al. \cite{nayel2019nayel} investigated age and gender prediction on social networks, emphasizing the challenge of short text lengths (average of 12 words). They utilized emoticons, image sequences, and character/word n-grams as features, achieving optimal results with a balanced training dataset. Parres et al. \cite{parres2022bert}
introduced diverse features, including structural metrics, part-of-speech analysis, emoticons, and themes derived from Latent Semantic Analysis (LSA). A Random Forest classifier with 311 features for age and 476 features for gender prediction yielded the highest accuracy. However, this approach was relatively slow.

\stitle{Learning Based:} 
Techniques primarily utilize supervised learning approaches. Hamda et al. \cite{nayel2019nayel} utilized Support Vector Machine (SVM) classifier for age, gender, and language variety identification in Arabic, achieving accuracy up to 95.97\% for age and 81.53\% for gender. In contrast, Schaetti et al. \cite{schaetti2017unine} combined TF/IDF with Convolutional Neural Networks (CNNs) to predict language variety and gender, showing higher accuracy rates of 98\% on Portuguese. Our model did not perform as well as theirs, even if it worked great for CNN and other neural networks.

\stitle{Corpus Based:}
Hsieh et al. \cite{hsieh2018author} investigated author profiling on \textit{Facebook}, achieving high accuracy in tasks such as age and gender prediction. Guimaraes et al. \cite{guimaraes2017age} explored age categorization on \textit{Twitter} corpus, achieving notable results using CNN. Zhao et al. \cite{zhao2017research} investigated age and gender identification in \textit{emails} and achieved accuracies of 72.10\% for gender and 81.15\% for age classification. Nguyen et al. \cite{nguyen2014gender} conducted a comparative analysis, highlighting limitations in current computational approaches. The study reveals that over 10\% of \textit{Twitter} users do not conform to their biological sex, and older users are often mistaken for younger ones.

\stitle{Dataset Based:}
Datasets like PAN have been instrumental in hosting various author profiling tasks, exploring traits such as gender, age, native language, and personality. While SVM remains a popular choice among participants \cite{bayot2016author}\cite{rahman2021multi}, techniques like random forest and logistic regression have also been employed \cite{dichiu2016using}. Character-level CNNs and RNNs also demonstrated success, offering alternatives to word-based approaches \cite{bagnall2015author}. In the 2016 PAN task, Vollenbroek et al. \cite{op2016gronup} achieved the highest average accuracy of 52.58\% using a linear SVM with n-grams and part-of-speech feature. However, Modaresi et al. found that part-of-speech tagging was not sufficient for domain-independent profiling \cite{modaresi2016exploring}. CLEF-PAN 2018 challenge \cite{stamatatos2018overview} included gender prediction from texts and photos, introducing new subtasks and language options.

\medskip

\stitle{\ex{Research Gap:}} The availability of a corpus is a need for developing any NLP technique in any language. The lack of an author dataset is the biggest challenge to authorship classification in the Bengali language. We created this dataset (BN-AuthProf) in order to address this problem. The dataset includes 30,131 manually collected Bangla Facebook posts with age and gender tags, along with 300 anonymized authors. Various classical machine learning and deep learning techniques were employed to evaluate the dataset by investigating optimized hyperparameters (such as n-gram, vectorizer, loss, activation, kernel size, batch size, and epoch).

\section{BN-A\lowercase{uth}P\lowercase{rof} Dataset}\label{sec:dataset}
In this section, we comprehensively explore the BN-AuthProf dataset, its creation, labeling process, and features.

\vspace{-1mm}
\subsection{\textbf{Data Accumulation}}

Social media platforms like Facebook, Twitter, and Instagram are widely utilized in both Bangladesh and India. Facebook, in particular, holds substantial prominence, with approximately 90.46\% usage in Bangladesh\footnote{\url{https://gs.statcounter.com/social-media-stats/all/bangladesh}}, and India leading globally in terms of Facebook users\footnote{\url{https://www.statista.com/statistics/268136/}}. Consequently, Facebook was chosen as the primary platform for data collection. Our author selection includes Facebook users from Bangladesh and West Bengal (an Indian province), spanning various professions. We manually collected textual data from 300 authors, each contributing numerous Bengali posts. We contacted each author before collecting their data and only chose those who gave consent, following ethical data collection practices. Personal information, such as user IDs, was systematically removed, and numerical IDs from 1 to 300 were assigned for data accumulation and labeling. We followed several guidelines during data collection:

\stitle{Author Selection:} We only selected authors having a substantial amount of posts ranging from 80 to 100 status.

\stitle{Content Criteria:} Status containing more than 10 words were included to ensure data quality in terms of semantics.

\stitle{Language Integrity:} Statuses with mixed languages were excluded to maintain focus on Bengali.

\stitle{Privacy:} Statuses containing personal information or sensitive data were avoided to maintain the privacy of authors.

\stitle{Exclusion of External Links:} Statuses containing URLs or external links were excluded to maintain data integrity.

\stitle{Data Deduplication:} Duplicated statuses from the same author were removed to ensure dataset uniqueness.

\stitle{Exclusion of Shared Content:} Only original posts from chosen authors were included, excluding shared content.


\subsection{\textbf{Data Annotation}}
We have classified authors based on two criteria: age and gender. Our dataset comprises 300 text documents (one per author), each named `\#User\_Id.txt', and a separate single file titled `\#Truth.txt'.  For example `1.txt' file contains the first authors' posts, with each line representing an individual post. Simultaneously, the corresponding labels (i.e., age, gender) are placed within the `\#Truth.txt' file, employing a specific format: `\texttt{\#1.txt:::<gender>:::<age>}'.

 \begin{figure}
    \centering
    \includegraphics[width=0.8\linewidth]{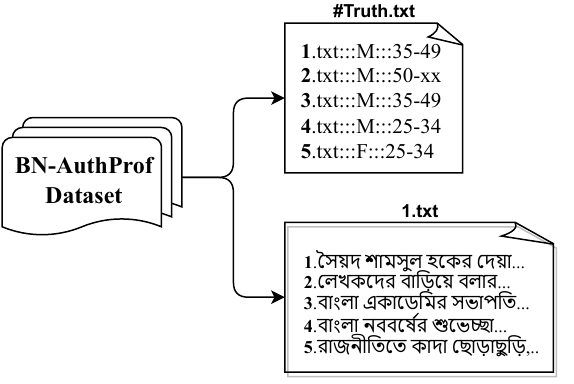}
    \vspace{-2mm}
    \caption{File Structure of BN-AuthProf Dataset}
    \label{fig:file-format}
    \vspace{-2mm}
  \end{figure}

In the context of the BN-AuthProf dataset, gender classification employs `\textbf{M}' to signify \textit{males} and `\textbf{F}' for \textit{females}. In terms of age classification, labels span four ranges:  `\textbf{18-24}', `\textbf{25-34}', `\textbf{35-49}', and `\textbf{50-xx}'. Fig. \ref{fig:file-format} provides a visual representation of our file structure.

\subsection{\textbf{Dataset Statistics}}
Table \ref{tab:Quantitative-Analysis-dataset} delineates the quantitative statistics of our dataset, where we observed a notable imbalance in gender and age groups, reflecting broader societal trends in Bangladesh\footnote{\url{https://datareportal.com/reports/digital-2023-bangladesh}}.

\vspace{-1mm}
\begin{table}[H]
\centering
\caption{Statistics of BN-AuthProf Dataset.}
\vspace{-2mm}
\label{tab:Quantitative-Analysis-dataset}
\begin{tabular}{p{4cm}l} 
\toprule
\textbf{Number of} & \textbf{Quantitative Value}\\ 
\midrule
Total Authors & 300\\ 
Status Per Author & 80 - 100\\ 
Total Status & 30131\\
Categories & 2 (Gender, Age)\\
\midrule
Gender: [Male (M), Female (F)] & [227, 73]\\
\midrule
Age: [18-24, 25-34, 35-49, 50-xx] & [43, 127, 82, 48]\\
\bottomrule
\end{tabular}
\end{table}

\subsection{\textbf{Data Augmentation}}
\begin{figure}[H]
    \centering
        \begin{tikzpicture}[scale = 0.55]
        \footnotesize
        \pie [polar]{75.7/Male,24.3/Female}
            
        \end{tikzpicture}
        \begin{tikzpicture}[scale = 0.55]
        \footnotesize
        \pie [polar]{42.3/25-34,
                    14.3/18-24,
                    27.3/35-49,
                    16/50-xx}
        \end{tikzpicture}
    \vspace{-1mm}
    \caption{Distribution of Gender and Age Categories}
    \label{fig:label-pie-chart}
\end{figure}
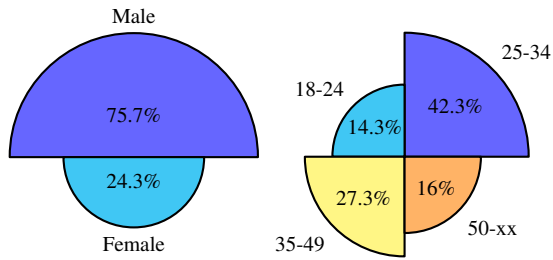

The pie charts in Fig. \ref{fig:label-pie-chart} illustrate the distribution of gender and age labels in our original dataset. There is a notable inconsistency in the distribution of these labels, which poses a challenge during benchmarking with machine learning algorithms by introducing bias toward the `M' gender label and the `25-34' age group. To address this issue, we have employed a solution that involves randomly generating additional data from the original dataset to produce a balanced training set.

\vspace{-2mm}

Initially, our dataset included data from 227 male and 73 female authors. To rectify this gender imbalance, we created 154 additional data entries for female authors. This augmentation was achieved by randomly selecting status updates from the existing female data and mixing them to create new, synthetic examples. Regarding age, we had 43 instances for the `18-24' group, 127 for the `25-34' group, 82 for the `35-49' group, and 48 for the `50-xx' group. To align these groups more closely with the dominant `25-34' group, we generated an additional 84 instances for the `18-24' group, 45 instances for the `35-49' group, and 79 instances for the `50-xx' group. The same random oversampling technique for generating gender-specific data was applied to balance the age groups. Table \ref{tab:Augmented_BN-AuthProf} provides insights into the adjustments made to address the initial imbalance in gender and age labels.

\vspace{-2mm}
\begin{table}[H]
\caption{Augmented Balanced Dataset after Oversampling}
\vspace{-2mm}
\label{tab:Augmented_BN-AuthProf}
\resizebox{\columnwidth}{!}{%
\begin{tabular}{lllllllllll}
\toprule
\multirow{2}{*}{\textbf{Dataset}} &  & \multicolumn{3}{c}{\textbf{Gender}} &  & \multicolumn{5}{c}{\textbf{Age}} \\ \cline{3-5} \cline{7-11} 
 &  & \multicolumn{1}{c}{\textbf{M}} & \multicolumn{1}{c}{\textbf{F}} & \multicolumn{1}{c}{\textbf{Total}} &  & \multicolumn{1}{c}{\textbf{18-24}} & \multicolumn{1}{c}{\textbf{25-34}} & \multicolumn{1}{c}{\textbf{35-49}} & \multicolumn{1}{c}{\textbf{50-xx}} & \multicolumn{1}{c}{\textbf{Total}} \\ \cline{1-1} \cline{3-5} \cline{7-11}
\textbf{Original} &  & 227 & 73 & 300 &  & 43 & 127 & 82 & 48 & 300 \\
\textbf{Augmented} &  & 0 & 154 & 154 &  & 84 & 0 & 45 & 79 & 208 \\
\textbf{Experimental} &  & 227 & 227 & 454 &  & 127 & 127 & 127 & 127 & 508 \\
\bottomrule
\end{tabular}%
}
\vspace{-2mm}
\end{table}

\subsection{\textbf{Experimental Dataset}} \label{sec:experimental-dataset-BN-AuthProf}
For benchmarking, we created the experimental dataset that combines original data with augmented data (see Table \ref{tab:Augmented_BN-AuthProf}). This experimental dataset is divided into two distinct categories: Gender and Age. We further split each category into training, development, and testing sets \textit{based on the number of posts}. We allocated approximately 80\% of each author's posts to training, 10\% to validation, and 10\% to the testing portion. Table \ref{tab:Experimental-Dataset-Gender} and Table \ref{tab:Experimental-Dataset-Age} present detailed statistics of the data. It is important to note that we only used the augmented data in the training set, while the validation and test sets consisted solely of original data. This decision ensures that the validation and test sets are not repeated during our evaluations.

\vspace{-2mm}
\begin{table}[H]
\caption{Experimental Dataset for Gender Classification Task}
\vspace{-2mm}
\label{tab:Experimental-Dataset-Gender}
\resizebox{\columnwidth}{!}{%
\begin{tabular}{llllllllll}
\hline
\textbf{Exp. Dataset} &  & \multicolumn{2}{c}{\textbf{Train}} &  & \multicolumn{2}{c}{\textbf{Valid}} &  & \multicolumn{2}{c}{\textbf{Test}} \\ \cline{3-4} \cline{6-7} \cline{9-10} 
\textbf{(Gender)} &  & \textbf{\#Doc} & \textbf{\#Posts} &  & \textbf{\#Doc} & \textbf{\#Posts} &  & \textbf{\#Doc} & \textbf{\#Posts} \\ \cline{1-1} \cline{3-4} \cline{6-7} \cline{9-10} 
\textbf{Male} &  & 277 & 18078 &  & 227 & 2256 &  & 227 & 2313 \\
\textbf{Female} &  & 277 & 17211 &  & 73 & 747 &  & 73 & 768 \\ \hline
\end{tabular}%
}
\vspace{-2mm}
\end{table}

\vspace{-3mm}
\begin{table}[H]
\caption{Experimental Dataset for Age Classification Task}
\vspace{-2mm}
\label{tab:Experimental-Dataset-Age}
\resizebox{\columnwidth}{!}{%
\begin{tabular}{llllllllll}
\hline
\textbf{Exp. Dataset} &  & \multicolumn{2}{c}{\textbf{Train}} &  & \multicolumn{2}{c}{\textbf{Valid}} &  & \multicolumn{2}{c}{\textbf{Test}} \\ \cline{3-4} \cline{6-7} \cline{9-10} 
\textbf{(Age)} &  & \textbf{\#Doc} & \textbf{\#Posts} &  & \textbf{\#Doc} & \textbf{\#Posts} &  & \textbf{\#Doc} & \textbf{\#Posts} \\ \cline{1-1} \cline{3-4} \cline{6-7} \cline{9-10} 
\textbf{18-24} &  & 127 & 6944 &  & 43 & 416 &  & 43 & 427 \\
\textbf{25-34} &  & 127 & 10137 &  & 127 & 1267 &  & 127 & 1300 \\
\textbf{35-49} &  & 127 & 10387 &  & 82 & 835 &  & 82 & 859 \\
\textbf{50-xx} &  & 127 & 7673 &  & 48 & 485 &  & 48 & 495 \\\hline
\end{tabular}%
}
\end{table}


\section{Benchmarking and Evaluation} \label{sec:benchmarking-methodology-BN-AuthProf}

\begin{figure}[]
    \centering
    \includegraphics[width=0.9\linewidth]{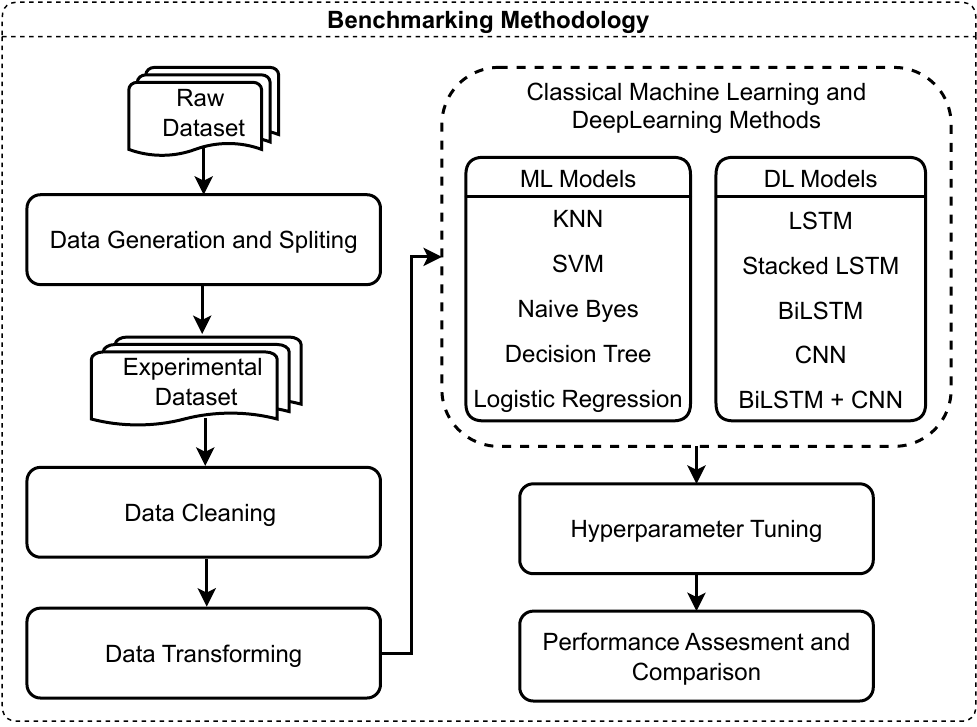}
    \vspace{-3mm}
    \caption{Benchmarking Approach for BN-AuthProf Dataset}
    \label{fig:EvaluationMethodology}
      \vspace{-2mm}
\end{figure}

For benchmarking, we used five classical ML models: Support Vector Machine, Naïve Bayes, Decision Tree, K-nearest Neighbor, Logistic Regression, and four deep learning models: LSTM, BiLSTM, CNN, BiLSTM + CNN. Fig. \ref{fig:EvaluationMethodology} outlines a structured approach for assessing the performance of the aforementioned models on the BN-AuthProf dataset. 

\subsection{Preprocessing}

\begin{figure}[H]
    \centering
    \includegraphics[width=0.9\linewidth]{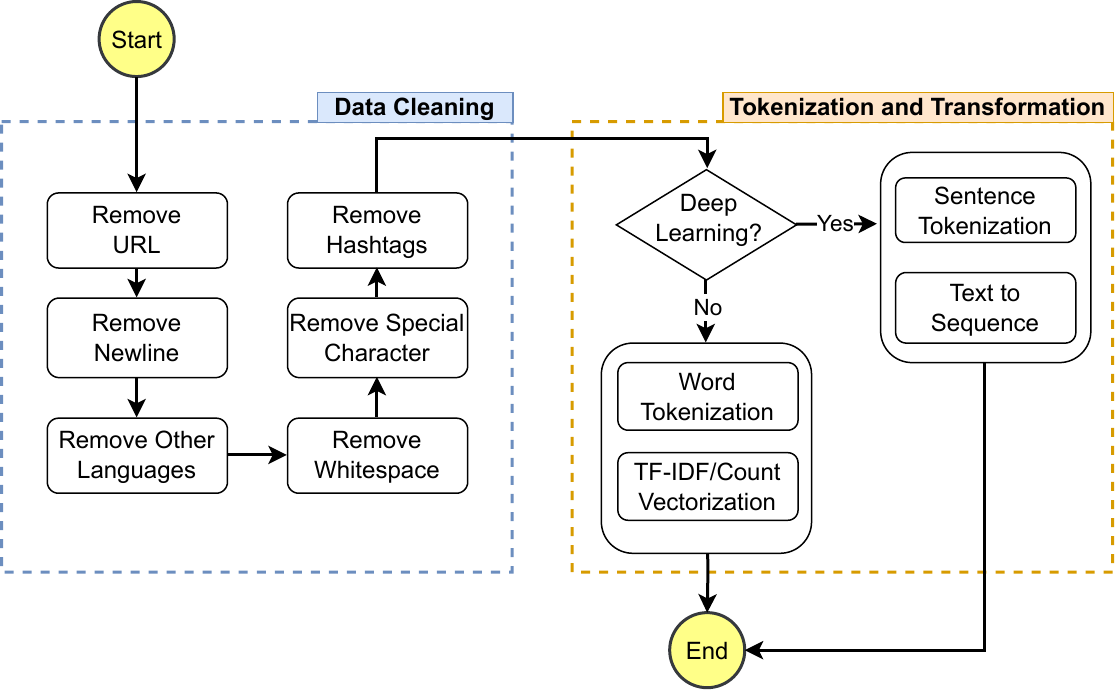}
    \vspace{-3mm}
    \caption{Data Preprocessing Pipeline}
    \label{fig:PreprocessingProcess}
    \vspace{-2mm}
\end{figure}

As we required only Bangla status updates, we had to clean irrelevant information from the original dataset. During the data preprocessing, we cleaned the data, tokenized it, and transformed the text into sequences. Fig. \ref{fig:PreprocessingProcess} shows a general overview of the data preprocessing pipeline with a flowchart.

\vspace{-2mm}

Briefly, we start by Removing URLs' from the data. Next, any `Line Feed' or `Newline Character' (i.e., $\backslash r$, $\backslash n$) is removed. After that, we remove any types of Latin Characters' that might constitute the English language. We also remove `Whitespaces', `Unwanted Characters', and the `Hashtag Symbol' since these elements do not contribute to the overall semantics of the data. However, we did not remove emojis, as they tend to polarize the status for certain groups, such as younger generations.

Since each status or post consists of multiple sentences, we need to chunk them into smaller units. This process is known as tokenization. Thus, in the next step, we determine whether we are applying classical machine learning approaches or deep learning techniques. For the deep learning models, we utilize `Sentence Tokenization', whereas for classical ones, we tokenize each input into words. For the deep learning approaches, we convert each sentence into a fixed-sized integer sequence (i.e., vector), where each element of the sequence represents either a word or padding. In contrast, for classical ML, the tokenized words are transformed into numerical representations using frequency-based techniques such as counting or TF-IDF (Term Frequency - Inverse Document Frequency).

\subsection{Classical Machine Learning}\label{subsubsec:MachineLearningModels}

The classical ML models are mostly statistical in nature and straightforward to utilize. However, they require handcrafted feature engineering, which offers greater insights into the models' behavior. Below, we provide details about the settings we used during our experiments with these classical models:

\stitle{Support Vector Machine:}
Support Vector Machine (SVM) is a supervised learning algorithm primarily used for classification tasks. It constructs an optimal decision boundary, or hyperplane, between data points of different classes. SVMs have been shown to excel in author profiling in the PAN shared task\cite{stamatatos2014overview}\cite{op2016gronup}\cite{stamatatos2018overview}. For gender classification, SVM employs binary classification using various kernel functions such as sigmoid, poly, and linear. Age classification utilizes multiclass techniques including One-vs-One (OvO) and One-vs-Rest (OvR), along with kernels like RBF and polynomials. We utilized both tf-idf and count vectorizers with character n-gram (range 1-8) and word n-gram (range 1-7) features.

\stitle{Naive Bayes:}
Naïve Bayes (NB) classifiers leverage Bayes' theorem to predict class labels based on feature independence assumptions. They are computationally efficient, making them ideal for rapid model development and prediction. Multinomial, Bernoulli, and Complement NB models are employed during the benchmarking with similar settings for n-gram features as SVM.

\stitle{Logistic Regression:}
Logistic Regression (LR) is a statistical method used to predict binary outcomes by modeling the relationship between a dependent variable and one or more independent variables. The tunable parameters in this model include different solvers: saga, liblinear, newton-cg, and lbfgs.

\stitle{K-Nearest Neighbor:}
K-Nearest Neighbor (KNN) is a nonparametric, supervised learning algorithm used for both classification and regression tasks. It classifies data points based on the majority vote of their nearest neighbors, with proximity serving as the determining factor. To optimize the performance of KNN, various distance metrics such as Minkowski, Euclidean, and Manhattan, as well as neighbor values ranging from 1 to 5, are explored.

\stitle{Decision Tree:}
Decision Trees (DT) are tree-structured classifiers that recursively split data based on feature attributes. They offer a simple yet powerful method for classification and regression tasks. Criteria functions like Gini and entropy, along with splitting strategies like best and random, are employed to construct and prune decision trees.

\subsection{Deep Learning}\label{subsubsec:DeepLearningModels}
Deep learning is a subset of machine learning that employs multi-layered neural networks for prediction tasks. Instead of relying on manual feature engineering, various hyperparameters play a pivotal role in performance tuning. Here, we will provide details about the hyperparameter settings we utilized during our experiments:

\stitle{LSTM and BiLSTM:}
Long Short-Term Memory (LSTM), a type of recurrent neural network (RNN), excels at handling sequential data by capturing long-term dependencies. In contrast, Bidirectional LSTM (BiLSTM) processes input in both directions, leveraging two layers of 128 LSTM cells each. We set varying activations, batch sizes, and epochs to optimize the performance of these models.

\stitle{Convolutional Neural Network:}
Convolutional Neural Networks (CNN) process sequential data using embedding layers, with the model architecture featuring convolutional layers with max-pooling and sigmoid activation. Trained with varying batch sizes and epochs, CNN is effective in feature extraction.

\stitle{BiLSTM+CNN:}
The combined model integrates BiLSTM and CNN, leveraging their complementary strengths. Employing the 'adam' optimizer and different loss functions, this model is trained with varying batch sizes and epochs. All deep learning models shared the same hyperparameters including maximum length, vocabulary size, and embedding dimensions, trained with adam optimizer,a learning rate of 0.001, and 16 samples per batch for 3, 5, 10, and 15 epochs.



\subsection{\textbf{Evaluation Techniques}}\label{sec:ClassificationTechniques}

\stitle{Measures of Classification Performance:} We evaluate the performance of different models on the experimental datasets (see Tables~\ref{tab:Experimental-Dataset-Gender} and \ref{tab:Experimental-Dataset-Age}). For task-specific supervision, we use the training data, while the validation set plays a pivotal role in tuning hyperparameters and optimizing the model. We perform evaluations on our test set using the best settings obtained during validation. We use \textbf{\textit{Precision}}, \textbf{\textit{Recall}}, \textbf{\textit{F1-score}}, and \textbf{\textit{Accuracy}} as evaluation measures.

\stitle{K-fold Cross Validation:} We also employed K-fold cross-validation to account for the model's generalizability. We used \(k=10\), dividing the whole dataset into 10 distinct folds for iterative processing to obtain a mean \textbf{Accuracy} score.

\subsection{\textbf{Best Performance}}\label{sec:best-performance-BN-AuthProf}
In our study, we conducted an extensive evaluation of numerous combinations of base classifiers for both gender and age prediction tasks. Among the classical models for \textbf{gender prediction}, the SVM model achieved the highest accuracy of \textit{80.6\%} and F1-score of \textit{0.614}. This result was achieved using an SVM with a polynomial kernel, word n-gram of (2, 3), and Tf-Idf feature vectorization. The Multinomial Naïve Bayes (MNB) proved to be a robust choice for gender prediction, delivering an accuracy of \textit{79.6\%} and f1-score of \textit{0.756}. MNB was particularly effective when accompanied by character n-gram of (2, 4) and Count feature vectorization.

Regarding \textbf{age prediction}, the Multinomial NB model outperformed others with an impressive accuracy of \textit{91\%} and F1-score of \textit{0.905}. This remarkable accuracy was achieved using word n-gram of (1, 8) and Count feature vectorization. Additionally, the SVM model, equipped with a polynomial kernel plus word n-gram of (2, 5), and Tf-Idf Vectorizer, achieved a competitive accuracy of \textit{66\%} and f1-score of \textit{0.582}. The results are summarized in Table \ref{tab:ResultsOverview}, showcasing the highest accuracy and f1-score attained by the two best models for gender and age classification.

\vspace{-2mm}
\begin{table}[H]
\centering
 \caption{Results of the Two Best Performed Models}
 \label{tab:ResultsOverview}
   \begin{tabular}{llll}
    \toprule
    \multicolumn{1}{l}{\textbf{Class}}& \textbf{Model} & \textbf{Accuracy} & \textbf{F1} \\
    \midrule
Gender   & SVM   & \ex{\textbf{0.806}}     & 0.614     \\
& NB    & 0.796     & \ex{\textbf{0.756}}     \\ \midrule
Age      & SVM   & 0.660     & 0.582     \\
& NB    & \ex{\textbf{0.910}}     & \ex{\textbf{0.905}} \\ 
\bottomrule
\end{tabular}
\end{table}

\subsection{\textbf{Comprehensive Analysis}}

\begin{figure*}
  \begin{minipage}[t]{.33\linewidth}
    \includegraphics[width=\linewidth]{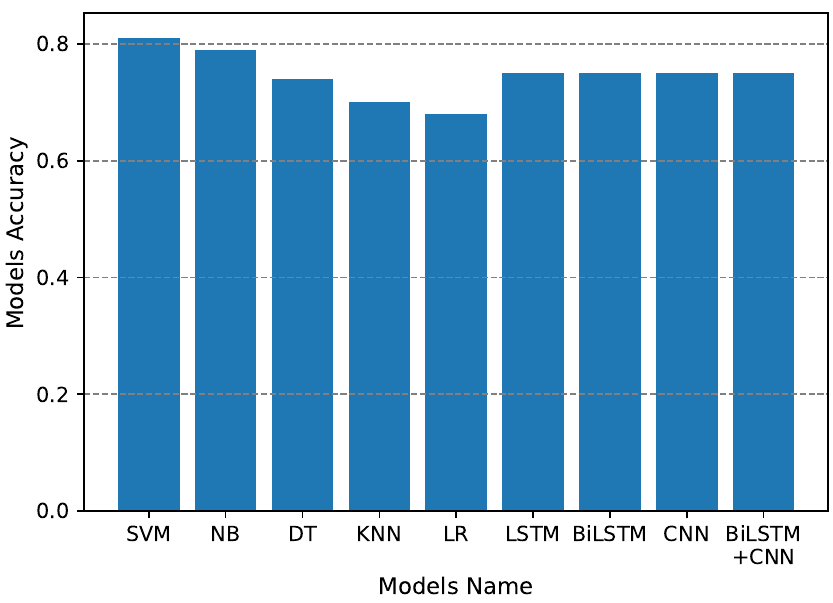}%
    \caption{Gender Classification Results}%
    \label{fig:AccuracyHistogramOfGender}
    \vspace{-2mm}
  \end{minipage}
  \hfil
    \begin{minipage}[t]{.31\linewidth}
    \includegraphics[width=\linewidth]{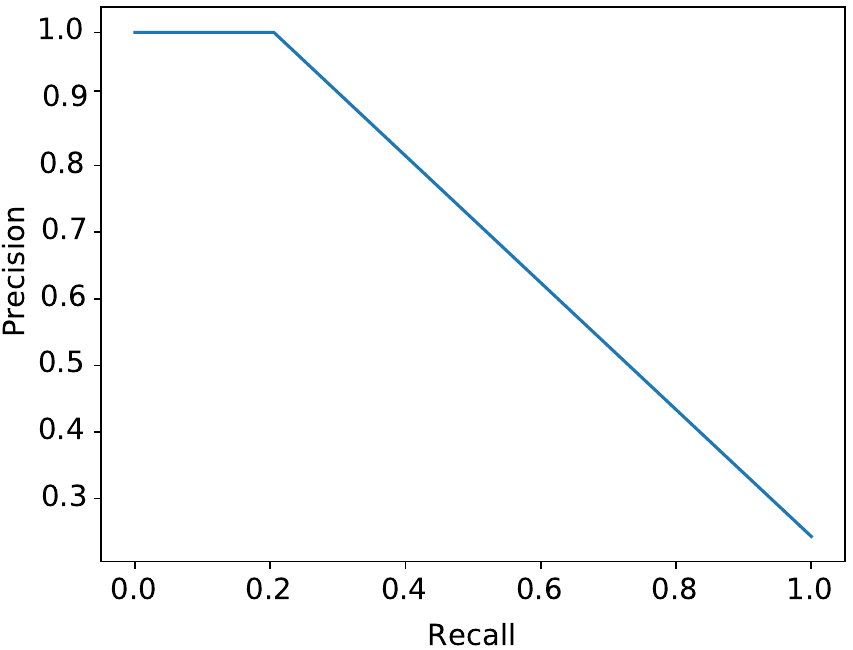}%
    \caption{Precision Vs. Recall Curve for SVM}%
    \label{fig:PrecisionRecallcurveforLogisticRegression}
    \vspace{-2mm}
  \end{minipage}
  \hfil
  \begin{minipage}[t]{.3\linewidth}
    \includegraphics[width=\linewidth]{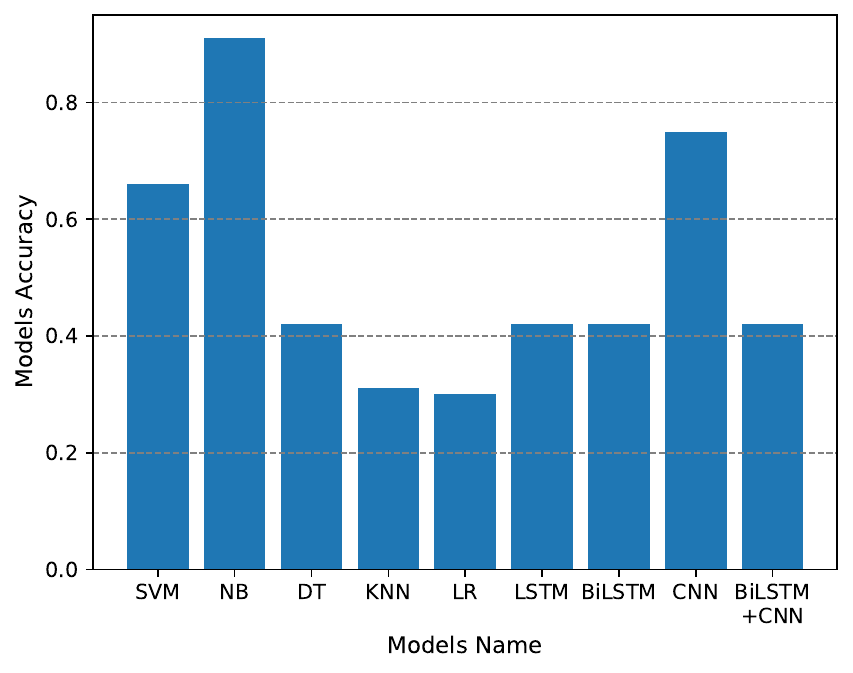}%
    \caption{Age Classification Results}%
    \label{fig:AccuracyMachineModelsAge}
    \vspace{-2mm}
  \end{minipage}
  \vspace{-2mm}
\end{figure*}

\stitle{\ex{Gender Classification:}} In this section, we present a thorough assessment of our gender classification experiments. We evaluated the performance of both classical machine learning and deep learning models, employing various feature extraction techniques, kernel functions, and vectorization methods. To commence, the bar chart of Fig. \ref{fig:AccuracyHistogramOfGender} showcases the distribution of accuracy of the benchmarked models. Remarkably, the Support Vector Machine (SVM) employing a word analyzer with a polynomial kernel, (2, 3) n-gram, and Tf-Idf Vectorizer achieved the highest accuracy of 80.6\%. The best combinations of hyperparameters achieved 75.67\% accuracy for deep learning models in gender prediction (see Table \ref{tab:ResultsDeepLearningMethodForGenderAndAge}).

Table \ref{tab:HighestResultMachineLearning} provides a comprehensive overview of the gender classification results, based on precision, recall, and F1-score scores. Notably, the SVM algorithm exhibited the highest precision score of 0.898, whereas the NB performed well in recall, achieving a score of 0.796. 


\begin{table}[H]
\centering
 \caption{Gender Classification Results of Classical ML}
 \vspace{-2mm}
 \label{tab:HighestResultMachineLearning}
\begin{tabular}{ccccc} 
\toprule
\textbf{Models} & \textbf{Accuracy} & \textbf{Precision}    & \textbf{Recall}   & \textbf{F1-Score}\\ 
\midrule
SVM& \ex{\textbf{0.806}}    & \ex{\textbf{0.898}}   & 0.603 & 0.614\\ 
NB& 0.796& 0.741& \ex{\textbf{0.796}}& \ex{\textbf{0.756}}\\ 
DT& 0.740& 0.489& 0.498& 0.449\\ 
LR& 0.683& 0.378& 0.496& 0.429\\ 
KNN& 0.700& 0.468& 0.486& 0.460\\
\bottomrule
\end{tabular}
\end{table}

We employed 10-fold cross-validation to assess the robustness of our system. Table \ref{tab:CrossValidationScoreForGender} presents the mean accuracy scores for the classical models. Support Vector Machine (SVM) emerged as the most accurate model, with an accuracy of 75.6\%, while Decision Tree (DT) exhibited the lowest accuracy at 74\%. Interestingly, NB consistently performed well, whereas K-Nearest Neighbor (KNN) and Logistic Regression (LR) showed the same accuracy.

\vspace{-2mm}
\begin{table}[!htb]
    \centering
     \caption{Cross Validation for Gender Prediction}
     \vspace{-1mm}
    \label{tab:CrossValidationScoreForGender}
    \begin{tabular}{cccccc}
    \toprule
     \textbf{Models} & \textbf{SVM} & \textbf{NB} & \textbf{DT} & \textbf{KNN} & \textbf{LR}\\
    \midrule
     Accuracy & \ex{\textbf{0.756}}& 0.750& 0.74& 0.753& 0.753\\ 
    \bottomrule
    \end{tabular}
\end{table}

\begin{figure}[h]
    \centering
    \vspace{-2mm}
    \begin{minipage}{0.50\linewidth}
        \centering
        \includegraphics[width=\linewidth]{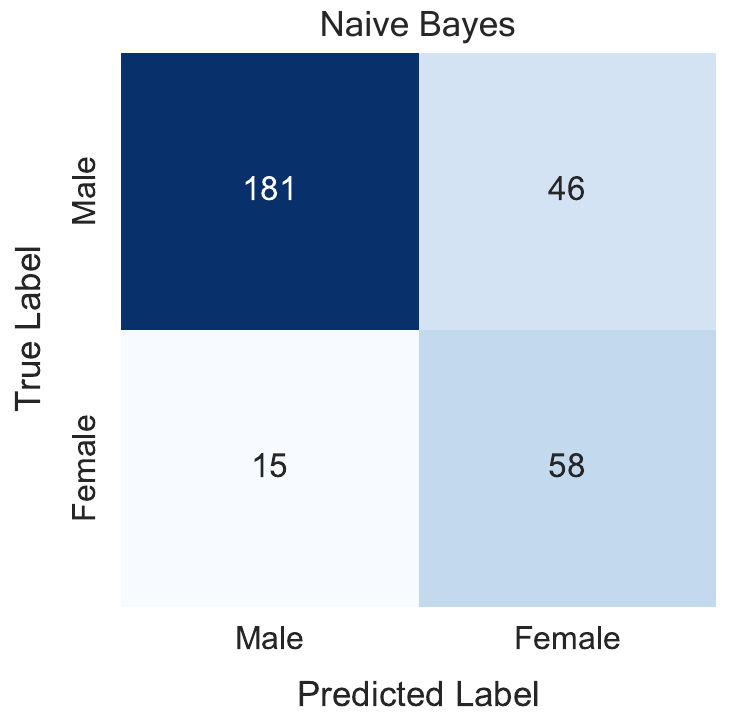}
    \end{minipage}\hfill
    \begin{minipage}{0.48\linewidth}
        \centering
        \includegraphics[width=\linewidth]{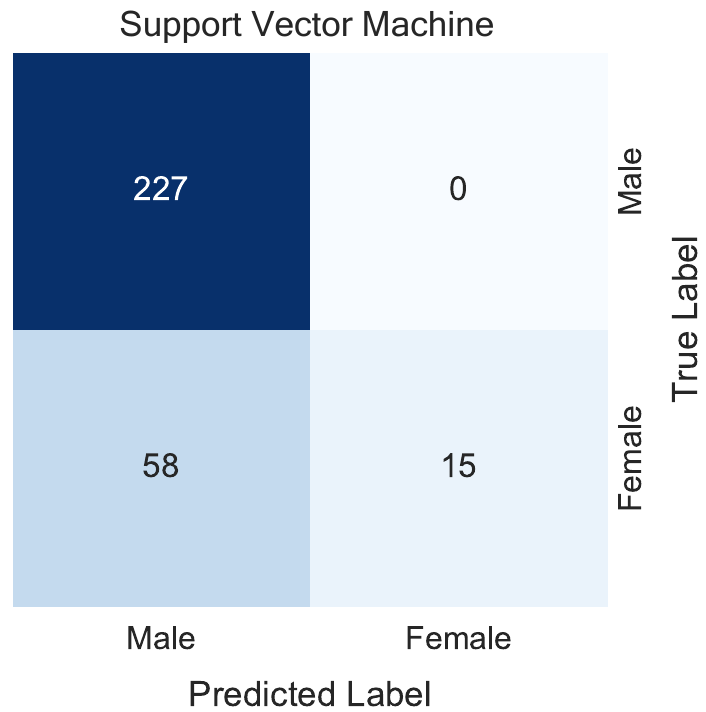}
    \end{minipage}
    \caption{Confusion Matrix for Gender Classification (Best)}
    \label{fig:time-vs-f1-vs-size-gender}
    \vspace{-2mm}
\end{figure}

To investigate further the performance of SVM and NB, we present the confusion matrices in Fig. \ref{fig:time-vs-f1-vs-size-gender}. It is evident that NB excelled in classifying females, accurately identifying 58 instances. However, SVM displayed a notable bias towards classifying instances as males, whereas NB demonstrates a balanced performance in both gender categories. In order to examine this phenomenon, we present the precision versus recall curve on varying classification margins (C values) of SVM in Fig. \ref{fig:PrecisionRecallcurveforLogisticRegression}. Examination of this curve reveals a gradual decline in precision once recall exceeds approximately 0.2. Accordingly, we established a threshold value of 0.67 for C, where precision is notably high while recall remains comparatively lower for the SVM classifier.

Turning to deep learning-based methods, Table \ref{tab:ResultsDeepLearningMethodForGenderAndAge} summarizes the results using the `adam’ optimizer in all cases. BiLSTM and CNN achieved the highest accuracy of 75.67\% when using the `binary crossentropy’ loss and `mean absolute percentage error’ loss functions. Both LSTM and CNN exhibited the lowest results with `mean absolute error loss’.

\stitle{\ex{Age Classification:}}
Age classification is a multi-class classification task, and we compared the models based on their overall average prediction scores among the classes. Fig. \ref{fig:AccuracyMachineModelsAge} provides a histogram of accuracy for our machine learning and deep learning models on age classification. Here, among all the models, Multinomial Naïve Bayes (MNB) achieved an impressive accuracy score exceeding 91\%. However, the overall average accuracy of the remaining models spans from as low as 31\% to as high as 71\%.


\vspace{-2mm}
\begin{table}[H]
\centering
\caption{Age Classification Results of Classical ML}
\vspace{-1mm}
\label{tab:ResultsMachineMethodonAgePrediction}
\begin{tabular}{ccccc} 
\toprule
\textbf{Models} & \textbf{Accuracy} & \textbf{Precision}    & \textbf{Recall}   & \textbf{F1-Score}\\
\midrule
SVM& 0.660& 0.889& 0.534& 0.582\\ 
NB&  \ex{\textbf{0.910}}& \ex{\textbf{0.892}}& \ex{\textbf{0.932}}& \ex{\textbf{0.905}}\\ 
DT& 0.420& 0.168& 0.249& 0.154\\ 
LR& 0.306& 0.236& 0.201& 0.160\\ 
KNN& 0.310& 0.276& 0.300& 0.277\\
\bottomrule
\end{tabular}
\vspace{-4mm}
\end{table}

\noindent Table \ref{tab:ResultsMachineMethodonAgePrediction} lists the results of age classification for the classical models. We observe that the NB algorithm outperformed all other models with the highest accuracy score of 0.91\% with similar precision and recall scores. SVM also demonstrated decent performance, with a precision of 0.88 and an F1 of 0.58. Table \ref{tab:CrossValidationScoreForAge} provides 10-fold cross-validation scores for various classical models. SVM in particular had the highest score of 42.3\%, whereas KNN and DT had the lowest accuracy.

\vspace{-3mm}
\begin{table}[H]
    \centering
     \caption{Cross Validation for Age Prediction}
     \vspace{-2mm}
     \label{tab:CrossValidationScoreForAge}
    \begin{tabular}{cccccc}
    \toprule
     \textbf{Models} & \textbf{SVM} & \textbf{NB} & \textbf{DT} & \textbf{KNN} & \textbf{LR}\\
     \midrule
     Accuracy & \ex{\textbf{0.423}} & 0.416 & 0.399 & 0.336 & 0.346\\ 
    \bottomrule
    \end{tabular}
    \vspace{-3mm}
\end{table}

\begin{figure}[h]
    \centering
    \begin{minipage}{0.50\linewidth}
        \centering
        \includegraphics[width=\linewidth]{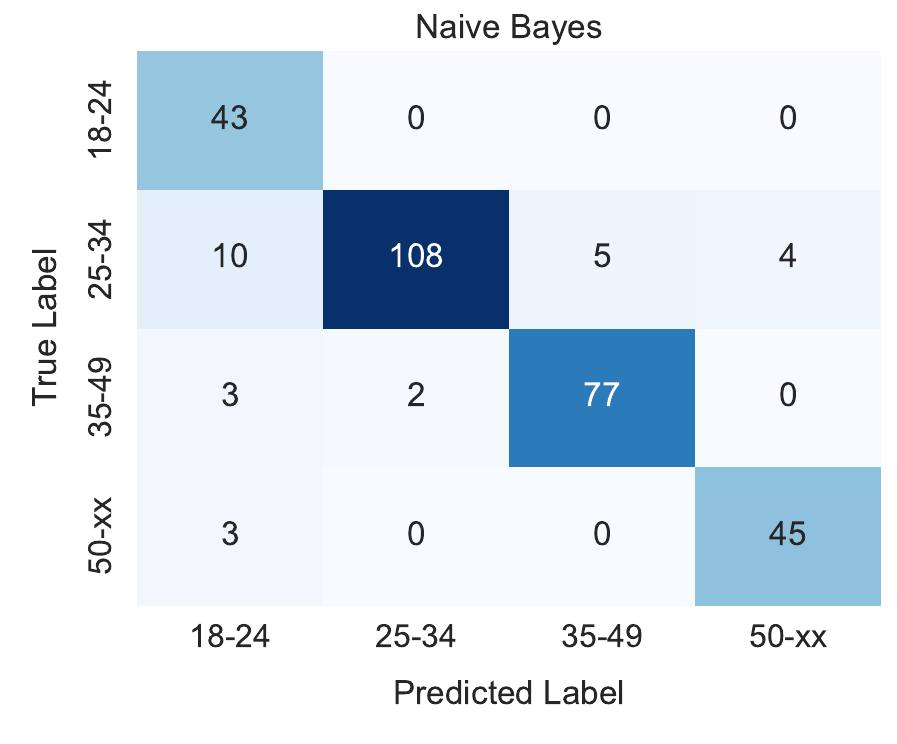}
    \end{minipage}\hfill
    \begin{minipage}{0.50\linewidth}
        \centering
        \includegraphics[width=\linewidth]{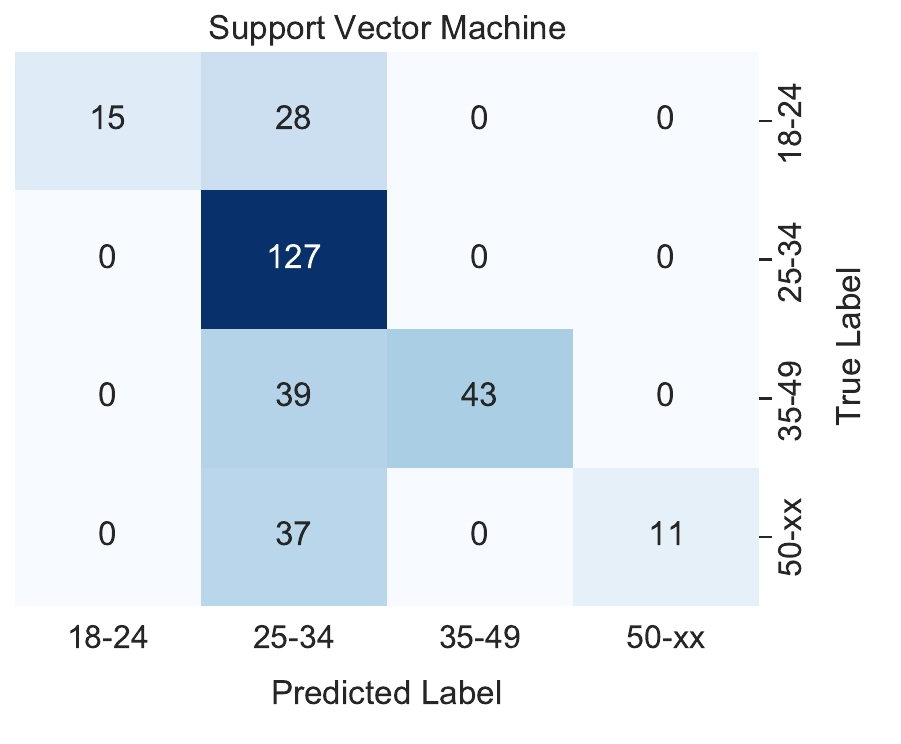}
    \end{minipage}
    \vspace{-2mm}
    \caption{Confusion Matrix for Age Classification (Best)}
    \vspace{-4mm}
    \label{fig:time-vs-f1-vs-size-age}
\end{figure}

We examined the confusion matrices for SVM and NB classifiers in Fig. \ref{fig:time-vs-f1-vs-size-age}. The SVM classifier demonstrates a distinct reliance on the 25–34 age class, showcasing notable proficiency in accurately categorizing instances within this particular age range. However, it is noteworthy that, within the 18-24 age class, only 15 instances were appropriately categorized, while the remaining 28 instances were misclassified and attributed to the 25-34 age class. Furthermore, SVM's performance displayed inconsistencies in accurately classifying instances within the 35-44 and 50-xx age classes, resulting in misclassifications. In contrast, NB exhibited a commendable level of accuracy across all age categories.

\vspace{-2mm}

Moving to deep learning models, exhibiting lower performance in general for age prediction. CNN stands out with the highest accuracy of 71\%. Meanwhile, LSTM and BiLSTM achieved accuracy scores ranging from 14\% to 42\% respectively. Table \ref{tab:ResultsDeepLearningMethodForGenderAndAge} provides an overview of deep learning-based age prediction using the `adam' optimizer in all cases. Interestingly, changing the activation function or optimizer did not significantly impact the scores.

\vspace{-2mm}
\begin{table}[H]
\centering
\caption{Deep Learning for Gender and Age Classification with Different Hyperparameters (Accuracy Scores).}
\vspace{-2mm}
\label{tab:ResultsDeepLearningMethodForGenderAndAge}
\resizebox{\columnwidth}{!}{%
\begin{tabular}{cccccc} 
\toprule
\textbf{Model} & \textbf{Activation} & \textbf{Loss} & \textbf{Batch} & \textbf{Gender} & \textbf{Age}\\ 
\midrule
LSTM& sigmoid& BFC  & 16& 0.75& 0.42\\ 
LSTM& softmax& MAE  & 32& 0.24& 0.42\\ 
BiLSTM& sigmoid& BC & 16& 0.75 & 0.42\\ 
BiLSTM& softmax& SCC    & 32& 0.75& 0.14\\ 
CNN& sigmoid& MAPE  & 16& \ex{\textbf{0.75}}& 0.14\\ 
CNN& relu& MAE  & 72& 0.24& \ex{\textbf{0.71}}\\ 
BiLSTM + CNN& sigmoid& MAE  & 16& 0.75& 0.42\\ 
BiLSTM + CNN& softmax& MAE  & 32& 0.24& 0.42\\
\bottomrule
\end{tabular}
}
\vspace{-2mm}
\end{table}
 
\stitle{\ex{Discussion:}} Our evaluation indicates that classical machine learning methods generally outperform deep learning in gender and age classification tasks. Naïve Bayes consistently emerged as a top performer. Additionally, SVM demonstrates strong performance. Specific model configurations and optimizers played a crucial role in DL models. These findings provide valuable insights into the application of machine learning algorithms for author profiling.

\section{Conclusions}
We have examined the field of author profiling for Bangla, extracting information about authors based on their writing style. This domain has been largely unexplored due to the unavailability of a dataset, despite having 300 million Bengali speakers worldwide. This paper presents the \textbf{BN-AuthProf} dataset encompassing 300 Bengali authors and 30,131 labeled Facebook status. Our investigation primarily revolved around crafting and evaluating machine learning and deep learning models for gender and age classification. Through rigorous experimentation, we attained promising benchmarks, particularly with Support Vector Machine and Naive Bayes algorithms. For our future work, we intend on expanding categories beyond age and gender and transitioning towards multi-label author profiling through transformer-based language models.

\bibliographystyle{IEEEtran}
\bibliography{references.bib}
\end{document}